\documentclass[a4paper,conference]{IEEEtran}
\usepackage{cite}

\ifCLASSINFOpdf
   \usepackage[pdftex]{graphicx}
\else
\fi
\usepackage{amsmath}
\usepackage{algorithmic}

\usepackage{array}
\usepackage{amsfonts}
\usepackage{mathtools}
\usepackage{diagbox}
\begin{document}
\title{ Learning Speaker-specific Lip-to-Speech Generation  }
 
\author{\IEEEauthorblockN{ \IEEEauthorrefmark{1} Munender Varshney\IEEEauthorrefmark{3} , \IEEEauthorrefmark{1} Ravindra Yadav\IEEEauthorrefmark{2}, Vinay P. Namboodiri\IEEEauthorrefmark{4} and Rajesh M Hegde\IEEEauthorrefmark{2}}
\IEEEauthorblockA{\IEEEauthorrefmark{2} Electrical department, Indian institute of Technology Kanpur, India \\}
\IEEEauthorblockA{\IEEEauthorrefmark{3} Computer Science and Engineering department, Indian institute of Technology Kanpur, India\\}
\IEEEauthorblockA{\IEEEauthorrefmark{4}University of Bath, UK \\
Email: munenderv@cse.iitk.ac.in, ravin@iitk.ac.in, vpn22@bath.ac.uk, rhegde@iitk.ac.in}
}

\maketitle
\def\thefootnote{*}\footnotetext{The first two authors contributed equally}
\begin{abstract}

Understanding the lip movement and inferring the speech from it is notoriously difficult for the common person. The task of accurate lip-reading gets help from various cues of the speaker and its contextual or environmental setting. Every speaker has a different accent and speaking style, which can be inferred from their visual and speech features. This work aims to understand the correlation/mapping between speech and the sequence of lip movement of individual speakers in an unconstrained and large vocabulary. We model the frame sequence as a distribution of features from the transformer in an auto-encoder setting and learn the embeddings jointly that exploits temporal properties of both audio and video. We learn temporal synchronization using deep metric learning, which guides the decoder to generate speech in sync with input lip movements. The predictive posterior thus gives us the generated speech in speaker speaking style. We have trained our model on the Grid and Lip2Wav Chemistry lecture dataset to evaluate single speaker natural speech generation tasks from lip movement in an unconstrained natural setting. Extensive evaluation using various qualitative and quantitative metrics with human evaluation also shows that our method outperforms on Lip2Wav Chemistry dataset (large vocabulary in an unconstrained setting) by a good margin across almost all evaluation metrics and marginally outperforms the state-of-the-art on GRID dataset.  

\end{abstract}


%
\IEEEpeerreviewmaketitle


\section{Introduction}
Recognizing the speech content of a speaker from their visual information alone, i.e., from the sequence of lip movement and facial expressions, is well studied in the research community and is popular as lip-reading. The task of lip-reading has a major challenge of word-level inherent ambiguity due to the presence of various homophones, i.e., the phonemes (distinct units of sound in a language), which are auditorily distinct while having almost all indistinguishable lip shapes sequences at the time of speaking. For example, if someone is uttering a phoneme \texttt{/b/} in a word (bark) then this can be easily confused with phoneme \texttt{/p/} in a similar word like park \cite{chung2017lip,ephrat2017vid2speech,goldschen1996rationale}. Similarly, there are a huge amount of words in the English language that cannot differentiate based on lip-reading alone \cite{iezzoni2004communicating,ebert1995communication}. Hence, even professional lip readers depend on multiple information streams such as the topic on which person is speaking, familiarity with the person, linguistic knowledge, head gestures, and facial expression. The lip-reading task is very perplexing, which is supported by the fact that a human can adequately understand speech across various accents and in a noisy environment. However, at the same time, they perform comparatively poorly on lip-reading tasks \cite{chung2016lip,assael2016lipnet}.   

Nevertheless, lip-reading has a variety of practical applications in speech transcription where audio is either missing or noisy. Some of the applications are long-range listening in surveillance video, efficient speech recovery in a noisy environment \cite{afouras2018conversation}, re-dubbing and transcribing the archival silent films, and a silent speech control system for resolving simultaneous speech from multi-talker \cite{sun2018lip}. The ``voice inpainting" \cite{zhou2019vision}  generate speech from lip movement to replace a corrupted speech segment. Lip-reading also has crucial applications in helping speech-impaired people, such as hearing aids and generation of voice for people who communicate using lip movements only as they are unable to generate voice(aphonia). The advent of deep neural network models \cite{krizhevsky2012imagenet,szegedy2015going,varshney2021optimizing,simonyan2014very,singh2020cooperative} and the availability of huge labeled datasets have helped in achieving a significant improvement in the task of the lip to text generation to solve a great variety of applications \cite{afouras2018deep,chung2016lip}. These datasets consist of a large vocabulary in an unconstrained environment with thousands of speakers. However, this conventional task needs data to be annotated by the human \cite{Rajat_2020_CVPR_Workshops}(i.e., transcribe speech into text format).
On the other hand, the lip to speech generation task does not require any annotations; hence it has drawn considerable attention as an alternative form of lip reading. Fundamentally, generating speech from a silent video of lip movement can be observed as a function or mapping from visemes to their corresponding phonemes. The mapping is learned by ``observing" the lip movement of the speaker for a prolonged period \cite{iezzoni2004communicating} while learning from a short clips of lip movement is challenging due to homophenes \cite{9747559}. Therefore, learning viseme-to-phoneme mapping only need long talking video clip of people without annotation to remove ambiguity using added visual context.

Inspired by various studies \cite{lewkowicz2012infants,iezzoni2004communicating}, professional lip readers and individuals can easily do lip-reading for people with frequent contact, i.e., an infant learns to speak by observing the lip movement of people around. Our focus in this work is the generation of speech from lip movements and contextual cues in the video of an individual in their speaking style. The generation is performed by just observing a particular person's speech patterns for a prolonged time period, not for any random speaker in the wild. Given the recent success of transformer models \cite{transformer} in natural language processing and its applications \cite{bert,albert,transformer_xl,roberta,sentence_bert,bart}, these models are applied in various problem domains of other modalities such as audio, text, image and video. In recent times, transformers have  exhibits better representational capability as compared to CNNs, thus various  design formulation of the transformers with variational auto-encoder models have been applied in various multi-modal domain applications such as music generation \cite{transformer_plus_vae5}, story generation \cite{transformer_plus_vae2} and sentiment analysis \cite{transformer_plus_vae3}. We use the transformer model to learn latent distributions from both input modalities jointly and the output embeddings are the sample drawn from these latent space. We used the deep metric learning-based approach proposed in \cite{zheng2021adversarial} to learn the temporal synchronization of lip movement and speech. We have tested our model on GRID and lip2wav chemistry lecture datasets. It outperforms all the previous state-of-the-art across various evaluation metrics on Lip2Wav Chemistry dataset, having a large vocabulary in an unconstrained setting while marginally outperforms on the GRID dataset.   

Section II discusses the recent developments and describes our approach and model architectures in the next section. The training approach and speech prediction are described in Sections V and VI. Finally, the detailed experimental evaluation is discussed in the next section and we conclude the work in Section VIII.

\begin{figure*}[htbp]
\centering
\includegraphics[width=\textwidth]{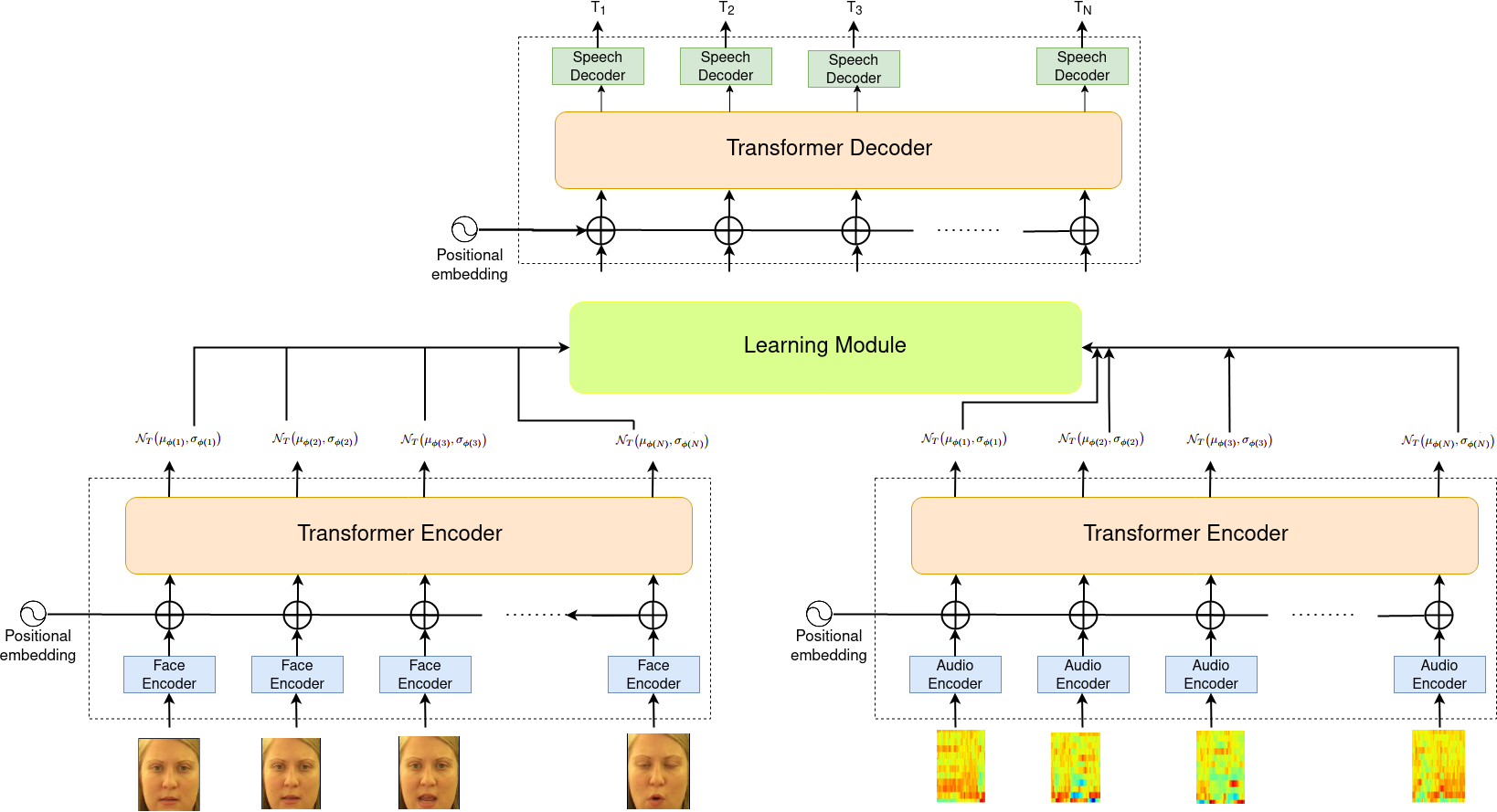}
\caption{\textbf{ Proposed model:} 
The network consists of individual video and audio transformers. In the learning module, the KL-divergence between the feature distribution of video networks and feature distribution of the audio network is minimized for each timestamp. At test time, the audio transformer decoder is used to predict the audio feature at time $t$ from the corresponding sample of the latent space. }
\label{proposed_model}
\end{figure*}

\section{Previous Work}

There is a vast history of work on lip reading; a nice detailed review of the work in the non-deep learning era is covered in the survey paper \cite{zhou2014review}. With advent of deep learning in the current era \cite{simonyan2014very,szegedy2015going} and the development of large-scale datasets \cite{krizhevsky2012imagenet} for lip-reading such as LRW \cite{chung2016lip}, LRS2 \cite{chung2017lip} and LRS3 \cite{afouras2018lrs3}, various work have used CNN to do recognition of smallest distinguishable units of sound from still images i.e., phonemes and visemes (visual equivalent of phonemes) \cite{noda2014lipreading,koller2015deep,chung2017lip}. The researchers have proposed various spatiotemporal end-to-end trainable CNN models such as ResNet and BiLSTM for word-level learning \cite{stafylakis2017combining,noda2014lipreading}. Next to this, the sentence level learning is performed using methodology of modelling automatic speech recognition(ASR) system using sequence-to-sequence models with LSTM or Connectionist Temporal Classification (CTC) \cite{graves2006connectionist,chung2017lip} i.e., Assael \emph{et. al}  proposed LipNet \cite{assael2016lipnet}, an end-to-end trainable model which uses CNN and LSTM-based architecture to map a variable-length sequence of frames to text using an auxiliary CTC loss for learning. The model achieves state-of-the-art performance in the speaker-independent setting of constrained grammar with vocabulary size ( 51 words) in GRID dataset \cite{cooke2006audio}. Other work in this direction uses the recently proposed powerful Transformer-based architectures \cite{afouras2018deep1} i.e., Conformer \cite{gulati2020conformer}, a hybrid architecture or its variants with different convolution blocks \cite{zhang2019spatio}. A few papers have explored the idea of knowledge distillation from ASR models, trained on the larger datasets to indirectly train the lip-reading model \cite{afouras2020asr,li2019improving}.

The ASR problem needs manual annotations in the form of text (transcription), while generation of speech, a closely linked problem, can be performed directly from lip movement. Preliminary research exploits CNN models to predict auditory features corresponding to a short video clip and replicate the same over a complete silent video using a sliding window to produce entire speech feature \cite{ephrat2017vid2speech,ephrat2017improved}. For example, Vid2Speech \cite{ephrat2017vid2speech} and Lipper \cite{kumar2019lipper} are trained end-to-end to predict LPC (Linear Predictive Coding) features from an input clip of $K$ frame using a 2D-CNN model and then these features are converted into speech. However, these features do not have significant speech information that leads to the generation of robotic speech and the model is inadequate to generate real-world talking face videos. The follow-up paper uses mel-spectrogram, a high-dimensional feature rather than LPC features, and the model uses optical flows with raw frames to enforce conditioning on lip motion. Akbari \emph{et al.} \cite{akbari2018lip2audspec} designed a lip-reading network having seven-layer 3D-CNN with LSTM to capture temporal information and use the generated feature to reconstruct auditory spectrogram from the decoder of the deep auto-encoder network. Researchers have also explored various generative approaches \cite{yadav2020bridged,yadav2020stochastic}, such as Vougioukas \emph{ et al.} \cite{vougioukas2019video} uses the 1D GAN model to synthesize raw waveform from a given video. Similarly, Michelsanti \emph{ et al.} \cite{michelsanti2020vocoder} uses a deep auto-encoder architecture where the encoder with a recursive module maps video frames of the speaker to vocoder features. There are five decoders for multitasking and the various outputs are used to reconstruct speech from the vocoder. 

Different from these work, Prajwal \emph{et al.} \cite{prajwal2020learning} gave a different perspective of mapping speech to lip sequence for an individual by adhering to their speaking style as well. They used a Seq2Seq architecture and a stack of 3D CNN network trained on a large vocabulary in an unconstrained setting. Yadav \emph{et al.} \cite{yadav2021speech} approaches speaker-specific generation using variational auto-encoder as a stochastic model and a recent work uses audiovisual analogy \cite{9747559}. Our work addresses the same problem of speech generation from the lip sequence of a specific speaker.     

From the trending success of Transformer models \cite{transformer} in natural language processing domain on various research problems and it's applications \cite{bert,albert,transformer_xl,roberta, sentence_bert,bart}, researchers have used these to solve the problem in other domains, such as image, audio, and video. Various transformers are proposed to handle a set of modalities such as video with text, image with text, and image with depth \cite{parida2022beyond}. These are famous as Multi-modal transformers \cite{gabeur2020multi,huang2020pixel}. These transformers can be used as a building block in a variational auto-encoder setting. This in turn help in generative modelling with multiple input modalities such as  text, audio, image and video \cite{transformer_plus_vae2,transformer_plus_vae5}. For example, Jiang \emph{et al.} designed a hierarchical model which unifies transformers for understanding long-term dependencies using their powerful attention mechanism and VAE to learn a disentangled latent space for music generation~\cite{transformer_plus_vae5}. On the same line of thought researchers has unified VAE with variant of transformers for various other applications such as  story generation~\cite{transformer_plus_vae2}, response generation~\cite{transformer_plus_vae4}, sentiment analysis~\cite{transformer_plus_vae3}, and 3D human pose generation~\cite{transformer_plus_vae1}. Recently, audio and visual modalities have been used jointly to improve various tasks such as zero-shot learning \cite{parida2020coordinated}, depth estimation \cite{parida2021beyond} etc.

\section{ Model architecture }

In this work, we have proposed the framework to integrate the latent variable model \cite{kingma2013auto} and the transformer \cite{transformer} to leverage the power of the generative model with the self-attention mechanism. Our model uses multi-modal data as input streams; one mode is a video stream, and the other is audio. The first input modality is video, i.e., a sequence of frames consisting of lip movement, which can be represented as $ \{ F_1, F_2, F_3 ..., F_t\}$ where $t$ represents the number of frames and their features are used as input to the video-transformer. At the same time, the other modality, speech, can be represented as $ \{ S_1, S_2, S_3, ..., S_t\}$ in Mel Frequency Cepstral Coefficients (MFCC ) feature space \cite{hossan2010novel} and given as input to the audio transformer. The initial output representation of both encoders is specific to modality. Our hierarchical framework uses transformer models as an abstraction to the variational auto-encoder Framework. The output features of both modalities from transformers are used to learn a joint cross-modal relationship between these embedding of the video and the audio sequence. The joint cross-modal relationship also exploits the temporal synchronization using deep metric learning \cite{zheng2021adversarial} and, at the same time, draws distributions of both modalities closer corresponding to the given timestamp. 

Our main objective is to learn speech prediction from the given sequence of silent video frames $ \{ F_1, F_2, F_3, ..., F_t\}$. We designed the framework to learn a decoder implicitly in the VAE framework using transformers as a building block. The speech prediction is performed using learned decoder (another transformer module) in the form of a sequence of audio window represented as $ \{ S_1, S_2, S_3, ..., S_t\}$. The generation of speech is govern by the transformer module  (deep generative network )  $ p_\theta (S | z)$, parameterized by $\theta $ and this decoder is defined by an auto-regressive form \cite{graves2013generating} 
\begin{equation}
p_\theta ( S | z ) = \prod_{t=1}^{T}p_{\theta }(S_t| S_{< t}, z)
\end{equation}
which depends on previously generated $S_{t-1}$`s to generate the next output MFCC feature \cite{hossan2010novel} element $S_{t}$. Thus in the proposed model, the decoder is a transformer due to its capability of effectively modeling the temporal relationships and the VAE framework \cite{kingma2013auto} enables the use latent space of $z$ to capture the distribution of the speech modality which helps generation. The proposed detailed  models is shown in the Fig \ref{proposed_model}. The VAE framework formed has a learnable distribution $p_{\phi_2} ( z | F)$ from the frame sequence $ \{ F_1, F_2, F_3, ..., F_t\}$  consist of lip movement. Training VAE need to maximize the marginal log-likelihood \cite{kingma2013auto} of audio data $ S $ for distribution $ \mathcal{D} $, which is given by 

\begin{equation}
\mathbb{E}_{S \sim \mathcal{D}} [{ \log p_{\phi_1, \phi_2, \theta }(S) }] 
\end{equation}

However, optimization of marginal likelihood is computationally infeasible. Consequently, we approximate the true posterior distribution $ \log p_{\phi_1} ( z| S )  \propto \log p_\theta (S|z) p(z) $ by introducing an $ \phi_1$-parameterized encoder which serve as an approximate inference model with distribution $\log q_{\phi_1} ( z| S ) $. The variational inference is used for training the VAE to yield the evidence lower bound (ELBO): 


\begin{equation}
\begin{aligned}
\mathbb{E}_{S \sim \mathcal{D}} \log p_{\{\phi_1,\phi_2, \theta, \}} ( S ) \geq  &
\mathbb{E}_{S \sim \mathcal{D}} \left [\mathbb{E}_{z \sim q_{\phi_1 }(z|S)}   \log p_\theta ( S |z ) \right ]  \\
& - \mathbb{E}_{S \sim \mathcal{D}}  [ \textit{ KL }  q_{\phi_1} (z|S) || p_{\phi_2} ( z | F ) )] \\
& \geq \mathcal{L}_{REC} - \mathcal{L}_{KL}
\end{aligned}
\label{eq:3}
\end{equation}
The first term in equation \ref{eq:3} is reconstruction loss and the second term is the KL Divergence between posteriors and the priors. 

\subsection{Video representation}

In the NLP transformer applied directly on the word tokens \cite{transformer}, while in the proposed model, we have frame sequence $ \{ F_1, F_2, F_3, ..., F_t\}$ therefore, the transformer operates on a sequence of embedding vectors obtained from the frame encoder network. The transformer is designed using a stack of $L$ layers of multi-head attention and a point-wise, fully connected feed-forward network corresponding to both encoder and decoder. The multi-head attention module has multiple self-attention networks working parallel to capture the temporal relationship and contextual clues in any given data sequence. In principle, a transformer maps the input into a query and key-value pairs. The corresponding attention is defined as a mapping between a query and the set of key-value pairs to give an output vector. The output vector is the weighted average of computed values and the corresponding weights is calculated using a compatibility function that operates between the query and the corresponding key, as shown below,

\begin{equation}
Attention(Q, K, V) = softmax(\frac{QK^T}{\sqrt{d_k}})V
\end{equation}   

Where Q represents a matrix consisting of a set of queries while variables K and V represent key and values, respectively, packed into a single matrix for all the inputs. These queries and keys of input have a dimension $d_k$, while values have a different dimension $d_v$. Thus, attention is computed by the dot-product of the query and the key vector. However, instead of computing attention only once, the transformer benefits by linearly projecting the queries, keys, and values multiple times in parallel ( called Multi-headed attention) and then projects the combined attention to yield final values. For a more exhaustive background description of Transformer models, we request readers to refer to~\cite{transformer, annotated_transformer}. Multi-head attention helps the model to utilize various representation sub-spaces at different positions to attend to their information jointly. 

Thus, the embedding vector sequence is given as input to the transformer. Instead of producing a single vector, the transformer model output is an isotropic Gaussian distribution of the latent space with parameters mean and variance as $ ( \mu_i^{f} , \sigma_i^{f}) $ for each $i^{th}$ time step. The output distribution of the model behaves as prior distribution $ p_{\phi_2} ( z | F ) $ for learning of posterior distribution jointly. We use these transformer models to learn the interaction between the samples obtained from these Gaussian distributions. 


\subsection{Audio representation }
First, we prepossess the given video to take out the audio waveform and calculate the MFCC features from these waveforms. The sequence of MFCC features thus obtained, can be represented as  $ \{ S_1, S_2, S_3, ..., S_t\}$ and given as input to transformer model instead of the raw audio. The multi-head attention mechanism of the transformer learns the temporal correlation between the sequence of MFCC features. In the given framework, the transformer learns a normal distribution as output for each timestamp, not just a single output vector. The parameters for learned distribution are  mean and variance $ ( \mu_i^{s} , \sigma_i^{s}) $. The learned Gaussian distribution represents the variational posterior of the VAE framework. Since the learnable prior is also Gaussian, the KL divergence is analytically solvable. At training time, a gradient is passed through Gaussian sampling using the re-parameterization trick \cite{kingma2013auto} to get the trained VAE model. 



\section{Training }

Training a network framework is equivalent to minimizing the loss function in equation \ref{eq:3}. Since, maximizing the marginal log-likelihood is intractable, so we maximize its evidence lower bound  ELBO \cite{kingma2013auto} which is equivalent to  minimizing the reconstruction loss and the KL divergence between posterior latent distribution $ q_{\phi_1} ( z|S) $ and the prior distribution $ p_{\phi_2} ( z | F ) $. Thus the distributions of both modalities come closer to each other for each timestamp, leading to a high correlation between the sequence of lip movement and its corresponding speech in the time domain. However, Learning the synchronization between speech and lip-movement sequence will also need the distributions of varying timestamps to be farther distant to help better speech generation. So, we can exploit deep metric learning to reduce the distance between the embeddings of the same timestamp (in synchronization ) and pull apart the embeddings of different timestamps in cross modalities.

\begin{figure}[htbp]
\centering
\includegraphics[width=8.59cm]{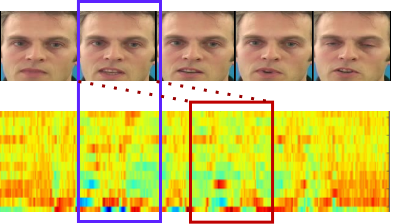}
\caption{ The blue marked face is the anchor  and it forms a positive pair with MFCC feature segment coloured in blue while anchor form a negative pair with  MFCC feature in red colour}
\label{metric}
\end{figure}

\subsection{ Deep metric learning  }
The focus of deep metric learning is to learn closer embedding for audio and video corresponding to same timestamp and a distant embedding for different timestamp. The audio-frame pair for same timestamp is termed as positive pair while for different timestamp is termed as negative pair and our objective is to preserve that implicit structure. Concretely, we select a frame $ F_i$ at random from the video, called anchor and select two audio feature sample, first is in-sync audio $S_i$ (positive pair) and other is out-off sync audio  sample $ S_j $ (negative pair) from the given video i.e., anyone expect $ S_i $. Finally, our deep metric learning approach uses a structured loss to pull the distance between the positive pair i.e. $F_i$ and $S_i$, and at the same time pushing the negative set ${S_j | j \neq i} $ away from the positive pair using the following learning objective function written below,

\begin{equation}
\mathcal{L}_{met} =\frac{1}{2N} \sum_{n=1}^{N} \max(0, \mathcal{J}_n)^2 
\label{eq:5}
\end{equation}
\begin{equation}
\mathcal{J}_n = \mathcal{D}(F_i, S_i) + \log ( e^{ \lambda - \mathcal{D}(F_i, S_j) } + e^{ \lambda - \mathcal{D}(S_i, S_j) }) 
\end{equation}

where $\mathcal{D}(F_i, S_j) $  represents the Euclidean distance between the $i^{th}$ frame feature of video $F$ and $j^{th}$ MFCC feature of speech $ S $. The hyper-parameter $ \lambda $ controls the distance margin between positive pair and negative set and $\mathcal{J}_n $ defines the similarity between positive pairs as well as dissimilarities of negative sets with the positive pairs. Therefore minimizing $\mathcal{J}_n $ helps in enlarging distances of out-off sync feature elements while minimizing the distance of in-sync feature elements.

\subsection{ Joint learning   }
Our model has integrate VAE with transformer and used metric learning for synchronisation. Thereof, we jointly learn modality-independent feature representation and synchronised feature embedding by minimizing the joint loss defined in equation \ref{eq:3} and \ref{eq:5}. The joint loss can be given as:
\begin{equation}
\mathcal{L}_{joint} = \mathcal{L}_{REC}  + \mathcal{L}_{KL} + \lambda \mathcal{L}_{met}  
\end{equation}
Where $\lambda$ is the hyper-parameter corresponding to metric learning loss, we choose the best value of $\lambda$ using the validation data.   

\section{Speech Prediction}
At test time, We need to predict the speech from the given stream of frame sequence $F={f1, f2, ..., fN } $ at hand only since we do not have the audio. Thus, at every timestamp $t$, transformer  encoder is used to obtain the posterior distribution $q(z|ft) $. The sample, thus obtained from the posterior distribution, is then passed into the audio decoder, a transformer network that operates on the latent space learned by the posterior distribution. Thus, the decoder transformer uses the modality-independent latent representation to generate the MFCC feature embedding in an auto-regressive manner. We maintain a small overlap across the sliding windows which reduces the boundary effect and the final waveform is obtained using the standard Griffin-Lim algorithm from MFCC features.




\section{ Experimental Results }
We have evaluated our proposed model for the speech prediction task on the two large scale publicly available datasets, GRID \cite{cooke2006audio} and Lip2Wav \cite{prajwal2020learning}. The GRID dataset has a limited vocabulary, where each speaker is speaking only six words chosen from the fixed dictionary. Therefore, to evaluate the performance on a dataset of much larger vocabulary size, we trained the proposed model on the Chemistry dataset introduced in Lip2wav \cite{prajwal2020learning}. The dataset consists of nearly 20 hours of video recording of online teaching; it has a vocabulary size of nearly 5000 words, which is much larger than the GRID dataset.

\subsection{Quantitative results}
For quantitative comparison, we used various metrics such as Perceptual evaluation of speech quality (PESQ)~\cite{pesq}, Extended short-time objective intelligibility (ESTOI)~\cite{estoi} and Standard Short-Time Objective Intelligibility (STOI)~\cite{stoi}. The STOI and ESTOI metrics measure the intelligibility of generated speech, while PESQ measures the speech quality by comparing it with the ground truth speech signal.

We report the values of these metrics for the Grid dataset in Table~\ref{tab:quantitative}. We observe that models that use the MFCC feature instead of the raw audio or LSP features perform comparatively better. We also observe that the proposed approach outperforms or is at par with the recent Lip2wav~\cite{prajwal2020learning} and~\cite{yadav2021speech} models on the given metrics. We have shared the generated samples on the project page link\footnote{Project page: https://sites.google.com/view/lip-to-speech/home} given below; we request the reader to listen to the generated results at their discretion.

\begin{table}[!t]
\renewcommand{\arraystretch}{1.3}
\caption{Results on GRID dataset}
\label{tab:quantitative}
\centering
\renewcommand{\arraystretch}{1.5}
\begin{tabular}{|c||c|c|c|}
\hline
Model &  STOI & ESTOI &  PESQ \\
\hline
\hline
Vid2Speech \cite{ephrat2017vid2speech} & 0.491 & 0.335 & 1.734\\
\hline
Lip2AudSpec \cite{akbari2018lip2audspec} & 0.513 & 0.352 & 1.673 \\
\hline
Konstantinos \emph{et al.} \cite{vougioukas2019video} & 0.564 & 0.361 & 1.684 \\
\hline
Ephrat \emph{et al.}~\cite{ephrat2017improved} & 0.659 & 0.376 & 1.825 \\
\hline
Lip2Wav~\cite{prajwal2020learning} & 0.731 & 0.535 & 1.772 \\
\hline
Yadav \emph{et al.}~\cite{yadav2021speech} & 0.724 & 0.540 & 1.932 \\
\hline
Proposed model & 0.711 & \textbf{0.543} & \textbf{2.022} \\
\hline
\end{tabular}
\end{table}

We report the results on the lip2wav chemistry lecture dataset in Table~\ref{tab:quantitative_chem}. These results show that the proposed framework outperforms the state-of-the-art Lip2wav approaches in both STOI and ESTOI metrics significantly ($0.416$ vs.\ $0.490$, $0.284$ vs.\ $0.316$) and performs marginally lower in the PESQ metric ($1.300$ vs.\ $1.233$).

\begin{table}[!t]
\renewcommand{\arraystretch}{1.3}
\caption{Results on Lip2wav Chemistry lecture dataset.}
\label{tab:quantitative_chem}
\centering
\renewcommand{\arraystretch}{1.5}
\begin{tabular}{|c||c|c|c|}
\hline
Model & STOI & ESTOI & PESQ \\
\hline
\hline
Konstantinos \emph{et al.}~\cite{vougioukas2019video} & 0.192 & 0.132 & 1.057 \\
\hline
Ephrat \emph{et al.}~\cite{ephrat2017improved} & 0.165 & 0.087 & 1.056 \\
\hline
Lip2Wav~\cite{prajwal2020learning} & 0.416 & 0.284 & 1.300 \\
\hline
Proposed model & \textbf{0.490} & \textbf{0.316} & 1.233 \\
\hline
\end{tabular}
\end{table}

\begin{figure*}[htbp]
\centering
\includegraphics[width=\textwidth]{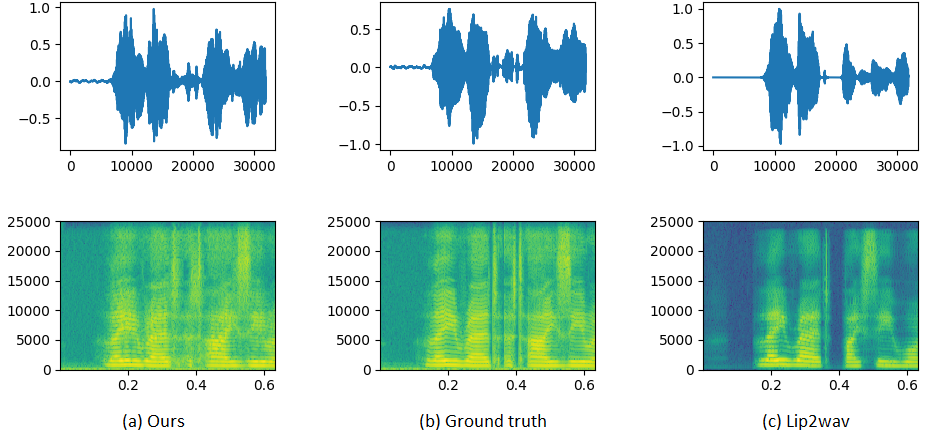}
\caption{\textbf{Qualitative comparison:} 
The figure compares the waveform and spectrogram outputs of proposed model and Lip2wav model with ground truth.}
\label{fig:spectrogram}
\end{figure*}

\subsection{Qualitative results}
We present the speech waveforms and the spectrogram generated by the Lip2wav model~\cite{prajwal2020learning} and the proposed model in Figure~\ref{fig:spectrogram}. These waveforms and the spectrogram indicate that the speech signal generated by the Lip2wav model is overly smoothed. Thus, the generated speech sounds robotic. On the other hand, the waveform and spectrograms generated by the proposed model match closely with the ground truth speech signal.

\subsection{User study}
We have done a subjective evaluation by conducting a user study. We asked ten subjects to rate the samples generated by the state-of-the-art models and our proposed framework model for the GRID dataset. Note that, in the GRID dataset, we have a fixed dictionary and each speaker speaks only six words taken from it. For each of the speech prediction models (~\cite{prajwal2020learning, yadav2021speech}, and ours), we showed randomly chosen 30 different prediction samples from a pool of 10 subjects. Thus, the samples shown to one subject may differ from those shown to another subject to counter the priming effect.

For each predicted speech sample, we asked the subjects to rate the speech quality based on how realistic (non-robotic) the generated speech sounds, i.e., the similarity of the generated speech to the ground truth for evaluating the speaker speaking style and correct pronunciation of words in sentences. All scores were represented on a scale of $1$ to $5$, where a higher score means better in for all three criteria and the results of corresponding studies are shown in Table ~\ref{tab:user_study}. We observe the predictions of our model are rated higher in all three criteria, which shows our results are more realistic and accurately understood (transcribed) by the listener.\\

\begin{table}[h]
\begin{center}
\caption{\textbf{User study}: Mean opinion score}
\label{tab:user_study}
\renewcommand{\arraystretch}{1.5}
\begin{tabular}{|c|c|c|c|}
\hline
\diagbox{Metric}{Model} & Yadav \emph{et al.} \cite{yadav2021speech} & Lip2Wav \cite{prajwal2020learning} & Ours \\
\hline
Similarity & 3.51 & 2.80 & 3.72 \\
\hline
Pronunciation & 3.42 & 2.98 & 3.50 \\
\hline
Realism & 3.58 & 2.95 & 3.65 \\
\hline
\end{tabular}
\end{center}
\end{table}

\section{Conclusion}
In this work, we look into synthesizing speech based on a sequence of lip movements from the perspective of individual speakers. We proposed a framework to unify the transformer model with VAE to learn a joint latent distribution space for smooth speech generation. We exploit the deep metric learning-based approach to learn the temporal synchronization of lip movement and speech. Our model outperforms state-of-the-art approaches on large-scale benchmark GRID and Lip2wav Chemistry datasets for the unconstrained, large vocabulary of single-speaker, which shows the effectiveness of our approach. We also observe from qualitative results and user studies that our approach produces more realistic speech than prior approaches.




\bibliographystyle{IEEEtran}
\bibliography{root}

\begin{thebibliography}{10}
\providecommand{\url}[1]{#1}
\csname url@samestyle\endcsname
\providecommand{\newblock}{\relax}
\providecommand{\bibinfo}[2]{#2}
\providecommand{\BIBentrySTDinterwordspacing}{\spaceskip=0pt\relax}
\providecommand{\BIBentryALTinterwordstretchfactor}{4}
\providecommand{\BIBentryALTinterwordspacing}{\spaceskip=\fontdimen2\font plus
\BIBentryALTinterwordstretchfactor\fontdimen3\font minus
  \fontdimen4\font\relax}
\providecommand{\BIBforeignlanguage}[2]{{%
\expandafter\ifx\csname l@#1\endcsname\relax
\typeout{** WARNING: IEEEtran.bst: No hyphenation pattern has been}%
\typeout{** loaded for the language `#1'. Using the pattern for}%
\typeout{** the default language instead.}%
\else
\language=\csname l@#1\endcsname
\fi
#2}}
\providecommand{\BIBdecl}{\relax}
\BIBdecl

\bibitem{chung2017lip}
J.~S. Chung, A.~Senior, O.~Vinyals, and A.~Zisserman, ``Lip reading sentences
  in the wild,'' in \emph{2017 IEEE Conference on Computer Vision and Pattern
  Recognition (CVPR)}.\hskip 1em plus 0.5em minus 0.4em\relax IEEE, 2017, pp.
  3444--3453.

\bibitem{ephrat2017vid2speech}
A.~Ephrat and S.~Peleg, ``Vid2speech: speech reconstruction from silent
  video,'' in \emph{2017 IEEE International Conference on Acoustics, Speech and
  Signal Processing (ICASSP)}.\hskip 1em plus 0.5em minus 0.4em\relax IEEE,
  2017, pp. 5095--5099.

\bibitem{goldschen1996rationale}
A.~J. Goldschen, O.~N. Garcia, and E.~D. Petajan, ``Rationale for
  phoneme-viseme mapping and feature selection in visual speech recognition,''
  in \emph{Speechreading by Humans and Machines}.\hskip 1em plus 0.5em minus
  0.4em\relax Springer, 1996, pp. 505--515.

\bibitem{iezzoni2004communicating}
L.~I. Iezzoni, B.~L. O'Day, M.~Killeen, and H.~Harker, ``Communicating about
  health care: observations from persons who are deaf or hard of hearing,''
  \emph{Annals of internal medicine}, vol. 140, no.~5, pp. 356--362, 2004.

\bibitem{ebert1995communication}
D.~A. Ebert and P.~S. Heckerling, ``Communication with deaf patients:
  knowledge, beliefs, and practices of physicians,'' \emph{Jama}, vol. 273,
  no.~3, pp. 227--229, 1995.

\bibitem{chung2016lip}
J.~S. Chung and A.~Zisserman, ``Lip reading in the wild,'' in \emph{Asian
  conference on computer vision}.\hskip 1em plus 0.5em minus 0.4em\relax
  Springer, 2016, pp. 87--103.

\bibitem{assael2016lipnet}
Y.~M. Assael, B.~Shillingford, S.~Whiteson, and N.~De~Freitas, ``Lipnet:
  End-to-end sentence-level lipreading,'' \emph{arXiv preprint
  arXiv:1611.01599}, 2016.

\bibitem{afouras2018conversation}
T.~Afouras, J.~S. Chung, and A.~Zisserman, ``The conversation: Deep
  audio-visual speech enhancement,'' \emph{arXiv preprint arXiv:1804.04121},
  2018.

\bibitem{sun2018lip}
K.~Sun, C.~Yu, W.~Shi, L.~Liu, and Y.~Shi, ``Lip-interact: Improving mobile
  device interaction with silent speech commands,'' in \emph{Proceedings of the
  31st Annual ACM Symposium on User Interface Software and Technology}, 2018,
  pp. 581--593.

\bibitem{zhou2019vision}
H.~Zhou, Z.~Liu, X.~Xu, P.~Luo, and X.~Wang, ``Vision-infused deep audio
  inpainting,'' in \emph{Proceedings of the IEEE/CVF International Conference
  on Computer Vision}, 2019, pp. 283--292.

\bibitem{krizhevsky2012imagenet}
A.~Krizhevsky, I.~Sutskever, and G.~E. Hinton, ``Imagenet classification with
  deep convolutional neural networks,'' \emph{Advances in neural information
  processing systems}, vol.~25, pp. 1097--1105, 2012.

\bibitem{szegedy2015going}
C.~Szegedy, W.~Liu, Y.~Jia, P.~Sermanet, S.~Reed, D.~Anguelov, D.~Erhan,
  V.~Vanhoucke, and A.~Rabinovich, ``Going deeper with convolutions,'' in
  \emph{Proceedings of the IEEE conference on computer vision and pattern
  recognition}, 2015, pp. 1--9.

\bibitem{varshney2021optimizing}
M.~Varshney and P.~Singh, ``Optimizing nonlinear activation function for
  convolutional neural networks,'' \emph{Signal, Image and Video Processing},
  vol.~15, no.~6, pp. 1323--1330, 2021.

\bibitem{simonyan2014very}
K.~Simonyan and A.~Zisserman, ``Very deep convolutional networks for
  large-scale image recognition,'' \emph{arXiv preprint arXiv:1409.1556}, 2014.

\bibitem{singh2020cooperative}
P.~Singh, M.~Varshney, and V.~Namboodiri, ``Cooperative initialization based
  deep neural network training,'' in \emph{Proceedings of the IEEE/CVF Winter
  Conference on Applications of Computer Vision}, 2020, pp. 1141--1150.

\bibitem{afouras2018deep}
T.~Afouras, J.~S. Chung, and A.~Zisserman, ``Deep lip reading: a comparison of
  models and an online application,'' \emph{arXiv preprint arXiv:1806.06053},
  2018.

\bibitem{Rajat_2020_CVPR_Workshops}
Rajat, M.~Varshney, P.~Singh, and V.~P. Namboodiri, ``Minimizing supervision in
  multi-label categorization,'' in \emph{Proceedings of the IEEE/CVF Conference
  on Computer Vision and Pattern Recognition (CVPR) Workshops}, June 2020.

\bibitem{9747559}
R.~Yadav, A.~Sardana, V.~P. Namboodiri, and R.~M. Hegde, ``Learning to predict
  speech in silent videos via audiovisual analogy,'' in \emph{ICASSP 2022 -
  2022 IEEE International Conference on Acoustics, Speech and Signal Processing
  (ICASSP)}, 2022, pp. 8042--8046.

\bibitem{lewkowicz2012infants}
D.~J. Lewkowicz and A.~M. Hansen-Tift, ``Infants deploy selective attention to
  the mouth of a talking face when learning speech,'' \emph{Proceedings of the
  National Academy of Sciences}, vol. 109, no.~5, pp. 1431--1436, 2012.

\bibitem{transformer}
A.~Vaswani, N.~Shazeer, N.~Parmar, J.~Uszkoreit, L.~Jones, A.~N. Gomez,
  {\L}.~Kaiser, and I.~Polosukhin, ``Attention is all you need,'' in
  \emph{Advances in neural information processing systems}, 2017, pp.
  5998--6008.

\bibitem{bert}
D.~Jacob, C.~Ming-Wei, L.~Kenton, and T.~Kristina, ``Bert: Pre-training of deep
  bidirectional transformers for language understanding,'' \emph{In NAACL},
  2019.

\bibitem{albert}
Z.~Lan, M.~Chen, G.~Sebastian, G.~Kevin, P.~Sharma, and S.~Radu, ``Albert: A
  lite bert for self-supervised learning of language representations,''
  \emph{In International Conference on Learning Representations}, 2020.

\bibitem{transformer_xl}
Z.~Dai, Z.~Yang, Y.~Yang, J.~Carbonell, Q.~V. Le, and R.~Salakhutdinov,
  ``Transformer-xl: Attentive language models beyond a fixed-length context,''
  \emph{In Association for Computational Linguistics}, 2019.

\bibitem{roberta}
Y.~Liu, M.~Ott, N.~Goyal, J.~Du, M.~Joshi, D.~Chen, O.~Levy, M.~Lewis,
  L.~Zettlemoyer, and V.~Stoyanov, ``Roberta: A robustly optimized bert
  pretraining approach,'' \emph{arXiv preprint arXiv:1907.11692}, 2019.

\bibitem{sentence_bert}
N.~Reimers and I.~Gurevych, ``Sentence-bert: Sentence embeddings using siamese
  bert-networks,'' \emph{In Association for Computational Linguistics}, 2019.

\bibitem{bart}
M.~Lewis, Y.~Liu, N.~Goyal, M.~Ghazvininejad, A.~Mohamed, O.~Levy, V.~Stoyanov,
  and L.~Zettlemoyer, ``Bart: Denoising sequence-to-sequence pre-training for
  natural language generation, translation, and comprehension,'' \emph{In
  Association for Computational Linguistics}, 2020.

\bibitem{transformer_plus_vae5}
J.~Jiang, X.~Gus~G., B.~C. Dave, A.~Chris~N., and M.~Ryan~H., ``Transformer
  vae: A hierarchical model for structure-aware and interpretable music
  representation learning,'' \emph{In ICASSP 2020-2020 IEEE International
  Conference on Acoustics, Speech and Signal Processing (ICASSP)}, pp.
  516--520, 2020.

\bibitem{transformer_plus_vae2}
L.~Fang, T.~Zeng, C.~Liu, L.~Bo, W.~Dong, and C.~Chen, ``Transformer-based
  conditional variational autoencoder for controllable story generation,''
  \emph{arXiv preprint arXiv:2101.00828}, 2021.

\bibitem{transformer_plus_vae3}
X.~Cheng, W.~Xu, T.~Wang, and W.~Chu, ``Variational semi-supervised aspect-term
  sentiment analysis via transformer,'' \emph{In Proceedings of the 23rd
  Conference on Computational Natural Language Learning (CoNLL)}, 2019.

\bibitem{zheng2021adversarial}
A.~Zheng, M.~Hu, B.~Jiang, Y.~Huang, Y.~Yan, and B.~Luo, ``Adversarial-metric
  learning for audio-visual cross-modal matching,'' \emph{IEEE Transactions on
  Multimedia}, 2021.

\bibitem{zhou2014review}
Z.~Zhou, G.~Zhao, X.~Hong, and M.~Pietik{\"a}inen, ``A review of recent
  advances in visual speech decoding,'' \emph{Image and vision computing},
  vol.~32, no.~9, pp. 590--605, 2014.

\bibitem{afouras2018lrs3}
T.~Afouras, J.~S. Chung, and A.~Zisserman, ``Lrs3-ted: a large-scale dataset
  for visual speech recognition,'' \emph{arXiv preprint arXiv:1809.00496},
  2018.

\bibitem{noda2014lipreading}
K.~Noda, Y.~Yamaguchi, K.~Nakadai, H.~G. Okuno, and T.~Ogata, ``Lipreading
  using convolutional neural network,'' in \emph{fifteenth annual conference of
  the international speech communication association}, 2014.

\bibitem{koller2015deep}
O.~Koller, H.~Ney, and R.~Bowden, ``Deep learning of mouth shapes for sign
  language,'' in \emph{Proceedings of the IEEE International Conference on
  Computer Vision Workshops}, 2015, pp. 85--91.

\bibitem{stafylakis2017combining}
T.~Stafylakis and G.~Tzimiropoulos, ``Combining residual networks with lstms
  for lipreading,'' \emph{arXiv preprint arXiv:1703.04105}, 2017.

\bibitem{graves2006connectionist}
A.~Graves, S.~Fern{\'a}ndez, F.~Gomez, and J.~Schmidhuber, ``Connectionist
  temporal classification: labelling unsegmented sequence data with recurrent
  neural networks,'' in \emph{Proceedings of the 23rd international conference
  on Machine learning}, 2006, pp. 369--376.

\bibitem{cooke2006audio}
M.~Cooke, J.~Barker, S.~Cunningham, and X.~Shao, ``An audio-visual corpus for
  speech perception and automatic speech recognition,'' \emph{The Journal of
  the Acoustical Society of America}, vol. 120, no.~5, pp. 2421--2424, 2006.

\bibitem{afouras2018deep1}
T.~Afouras, J.~S. Chung, A.~Senior, O.~Vinyals, and A.~Zisserman, ``Deep
  audio-visual speech recognition,'' \emph{IEEE transactions on pattern
  analysis and machine intelligence}, 2018.

\bibitem{gulati2020conformer}
A.~Gulati, J.~Qin, C.-C. Chiu, N.~Parmar, Y.~Zhang, J.~Yu, W.~Han, S.~Wang,
  Z.~Zhang, Y.~Wu \emph{et~al.}, ``Conformer: Convolution-augmented transformer
  for speech recognition,'' \emph{arXiv preprint arXiv:2005.08100}, 2020.

\bibitem{zhang2019spatio}
X.~Zhang, F.~Cheng, and S.~Wang, ``Spatio-temporal fusion based convolutional
  sequence learning for lip reading,'' in \emph{Proceedings of the IEEE/CVF
  International Conference on Computer Vision}, 2019, pp. 713--722.

\bibitem{afouras2020asr}
T.~Afouras, J.~S. Chung, and A.~Zisserman, ``Asr is all you need: Cross-modal
  distillation for lip reading,'' in \emph{ICASSP 2020-2020 IEEE International
  Conference on Acoustics, Speech and Signal Processing (ICASSP)}.\hskip 1em
  plus 0.5em minus 0.4em\relax IEEE, 2020, pp. 2143--2147.

\bibitem{li2019improving}
W.~Li, S.~Wang, M.~Lei, S.~M. Siniscalchi, and C.-H. Lee, ``Improving
  audio-visual speech recognition performance with cross-modal student-teacher
  training,'' in \emph{ICASSP 2019-2019 IEEE International Conference on
  Acoustics, Speech and Signal Processing (ICASSP)}.\hskip 1em plus 0.5em minus
  0.4em\relax IEEE, 2019, pp. 6560--6564.

\bibitem{ephrat2017improved}
A.~Ephrat, T.~Halperin, and S.~Peleg, ``Improved speech reconstruction from
  silent video,'' in \emph{Proceedings of the IEEE International Conference on
  Computer Vision Workshops}, 2017, pp. 455--462.

\bibitem{kumar2019lipper}
Y.~Kumar, R.~Jain, K.~M. Salik, R.~R. Shah, Y.~Yin, and R.~Zimmermann,
  ``Lipper: Synthesizing thy speech using multi-view lipreading,'' in
  \emph{Proceedings of the AAAI Conference on Artificial Intelligence},
  vol.~33, no.~01, 2019, pp. 2588--2595.

\bibitem{akbari2018lip2audspec}
H.~Akbari, H.~Arora, L.~Cao, and N.~Mesgarani, ``Lip2audspec: Speech
  reconstruction from silent lip movements video,'' in \emph{2018 IEEE
  International Conference on Acoustics, Speech and Signal Processing
  (ICASSP)}.\hskip 1em plus 0.5em minus 0.4em\relax IEEE, 2018, pp. 2516--2520.

\bibitem{yadav2020bridged}
R.~Yadav, A.~Sardana, V.~Namboodiri, and R.~M. Hegde, ``Bridged variational
  autoencoders for joint modeling of images and attributes,'' in
  \emph{Proceedings of the IEEE/CVF Winter Conference on Applications of
  Computer Vision}, 2020, pp. 1479--1487.

\bibitem{yadav2020stochastic}
R.~Yadav, A.~Sardana, V.~P. Namboodiri, and R.~M. Hegde, ``Stochastic talking
  face generation using latent distribution matching,'' in \emph{Proc.
  Interspeech 2020}, 2020.

\bibitem{vougioukas2019video}
K.~Vougioukas, P.~Ma, S.~Petridis, and M.~Pantic, ``Video-driven speech
  reconstruction using generative adversarial networks,'' \emph{arXiv preprint
  arXiv:1906.06301}, 2019.

\bibitem{michelsanti2020vocoder}
D.~Michelsanti, O.~Slizovskaia, G.~Haro, E.~G{\'o}mez, Z.-H. Tan, and
  J.~Jensen, ``Vocoder-based speech synthesis from silent videos,'' \emph{arXiv
  preprint arXiv:2004.02541}, 2020.

\bibitem{prajwal2020learning}
K.~Prajwal, R.~Mukhopadhyay, V.~P. Namboodiri, and C.~Jawahar, ``Learning
  individual speaking styles for accurate lip to speech synthesis,'' in
  \emph{Proceedings of the IEEE/CVF Conference on Computer Vision and Pattern
  Recognition}, 2020, pp. 13\,796--13\,805.

\bibitem{yadav2021speech}
R.~Yadav, A.~Sardana, V.~P. Namboodiri, and R.~M. Hegde, ``Speech prediction in
  silent videos using variational autoencoders,'' in \emph{ICASSP 2021-2021
  IEEE International Conference on Acoustics, Speech and Signal Processing
  (ICASSP)}.\hskip 1em plus 0.5em minus 0.4em\relax IEEE, 2021, pp. 7048--7052.

\bibitem{parida2022beyond}
K.~K. Parida, S.~Srivastava, and G.~Sharma, ``Beyond mono to binaural:
  Generating binaural audio from mono audio with depth and cross modal
  attention,'' in \emph{Proceedings of the IEEE/CVF Winter Conference on
  Applications of Computer Vision}, 2022, pp. 3347--3356.

\bibitem{gabeur2020multi}
V.~Gabeur, C.~Sun, K.~Alahari, and C.~Schmid, ``Multi-modal transformer for
  video retrieval,'' in \emph{Computer Vision--ECCV 2020: 16th European
  Conference, Glasgow, UK, August 23--28, 2020, Proceedings, Part IV 16}.\hskip
  1em plus 0.5em minus 0.4em\relax Springer, 2020, pp. 214--229.

\bibitem{huang2020pixel}
Z.~Huang, Z.~Zeng, B.~Liu, D.~Fu, and J.~Fu, ``Pixel-bert: Aligning image
  pixels with text by deep multi-modal transformers,'' \emph{arXiv preprint
  arXiv:2004.00849}, 2020.

\bibitem{transformer_plus_vae4}
Z.~Lin, W.~Genta~Indra, X.~Peng, L.~Zihan, and F.~Pascale, ``Variational
  transformers for diverse response generation,'' \emph{arXiv:2003.12738},
  2020.

\bibitem{transformer_plus_vae1}
M.~Petrovich, M.~J. Black, and G.~Varol, ``Action-conditioned 3d human motion
  synthesis with transformer vae,'' in \emph{Proceedings of the IEEE/CVF
  International Conference on Computer Vision}, 2021, pp. 10\,985--10\,995.

\bibitem{parida2020coordinated}
K.~Parida, N.~Matiyali, T.~Guha, and G.~Sharma, ``Coordinated joint multimodal
  embeddings for generalized audio-visual zero-shot classification and
  retrieval of videos,'' in \emph{Proceedings of the IEEE/CVF Winter Conference
  on Applications of Computer Vision}, 2020, pp. 3251--3260.

\bibitem{parida2021beyond}
K.~K. Parida, S.~Srivastava, and G.~Sharma, ``Beyond image to depth: Improving
  depth prediction using echoes,'' in \emph{Proceedings of the IEEE/CVF
  Conference on Computer Vision and Pattern Recognition}, 2021, pp. 8268--8277.

\bibitem{kingma2013auto}
D.~P. Kingma and M.~Welling, ``Auto-encoding variational bayes,'' \emph{arXiv
  preprint arXiv:1312.6114}, 2013.

\bibitem{hossan2010novel}
M.~A. Hossan, S.~Memon, and M.~A. Gregory, ``A novel approach for mfcc feature
  extraction,'' in \emph{2010 4th International Conference on Signal Processing
  and Communication Systems}.\hskip 1em plus 0.5em minus 0.4em\relax IEEE,
  2010, pp. 1--5.

\bibitem{graves2013generating}
A.~Graves, ``Generating sequences with recurrent neural networks,'' \emph{arXiv
  preprint arXiv:1308.0850}, 2013.

\bibitem{annotated_transformer}
\BIBentryALTinterwordspacing
A.~M. Rush, ``The annotated transformer,'' in \emph{Proceedings of workshop for
  NLP open source software (NLP-OSS)}, 2018, pp. 52--60. [Online]. Available:
  \url{https://nlp.seas.harvard.edu/2018/04/03/attention.html}
\BIBentrySTDinterwordspacing

\bibitem{pesq}
A.~Rix, J.~Beerends, M.~Hollier, and A.~Hekstra, ``Perceptual evaluation of
  speech quality (pesq)-a new method for speech quality assessment of telephone
  networks and codecs,'' \emph{In IEEE International Conference on Acoustics,
  Speech, and Signal Processing. Proceedings (Cat. No. 01CH37221). Vol. 2.
  IEEE}, p. 749–752, 2001.

\bibitem{estoi}
J.~Jensen and C.~H. Taal, ``An algorithm for predicting the intelligibility of
  speech masked by modulated noise maskers,'' \emph{In IEEE/ACM Transactions on
  Audio, Speech, and Language Processing 24.11}, pp. 2009--2022, 2016.

\bibitem{stoi}
C.~Taal, R.~Hendriks, R.~Heusdens, and J.~Jensen, ``A short-time objective
  intelligibility measure for time-frequency weighted noisy speech,'' \emph{In
  2010 IEEE international conference on acoustics, speech and signal
  processing}, pp. 4214--4217, 2010.

\end{thebibliography}


\end{document}